# Color Image Retrieval Using Fuzzy Measure Hamming and S-Tree

Thanh The Van[1,*], Thanh Manh Le[2]

[1] HoChiMinh city University of Food Industry, Vietnam
[2] Hue University, Vietnam

**Abstract.** This chapter approaches the image retrieval system on the base of the colors of image. It creates fuzzy signature to describe the color of image on color space HSV and builds fuzzy Hamming distance (FHD) to evaluate the similarity between the images. In order to reduce the storage space and speed up the search of similar images, it aims to create S-tree to store fuzzy signature relies on FHD and builds image retrieval algorithm on S-tree. Then, it provides the content-based image retrieval (CBIR) and an image retrieval method on FHD and S-tree. Last but not least, based on this theory, it also presents an application and experimental assessment of the process of querying similar image on the database system over 10,000 images.

**Keywords:** CBIR; Image Retrieval; FHD; Signature; S-Tree.

## 1 Introduction

It is difficult to find similar images in a large database of digital images. There is a solution to solve this problem, such as: text-based image retrieval (TBIR) based on the keywords [4, 5] but it is time-consuming and unfeasible for many different applications. Moreover, the process of labeling depends on the semantic description of the image. So the image retrieval system on the base of the content is developed to extract visual feature to describe the content of image. A number of image retrieval system was built as: QBIC, ADL, Virage, hAlta Vista, SIMPLYcity,…

In recent years, the works of query image regarding CBIR, such as: the image retrieval system on the base of the color histogram [4, 5], the similarity measure of image on the base of combining color and texture image [7, 8], image retrieval technique Variable-Bin Allocation (VBA) using signature and the S-tree [6],…

In the approach of the paper will create the fuzzy signature of an image. The content of the paper will aim to efficient query “similar images” in a large database system of digital image. There are two major targets are used to reduce the amount of storage space and speed up the query image on large database systems. It also builds an evaluation method of similarity measure of image on fuzzy measure Hamming and builds an image retrieval method on the S-tree.

[*] Thanh The Van (E-mail: thanhvt@cntp.edu.vn).

## 2 The Related Theory

### 2.1 Fuzzy Signature

The fuzzy signature $F$ with a length $m$ is a vector $(f_1, f_2, ..., f_m)$, with $f_i \in [0,1]$, $i = 1,...,m$ [1, 2].
The conjunction of fuzzy signatures $F^i$ and $F^j$ is a fuzzy signature:
$F^i \wedge F^j = (f_1^i \wedge f_1^j, f_2^i \wedge f_2^j, ..., f_m^i \wedge f_m^j)$, with $f_r^i \wedge f_r^j = \min\{f_r^i, f_r^j\}$, $r = 1,...,m$
The disjunction of fuzzy signatures $F^i$ and $F^j$ is a fuzzy signature:
$F^i \vee F^j = (f_1^i \vee f_1^j, f_2^i \vee f_2^j, ..., f_m^i \vee f_m^j)$, with $f_r^i \vee f_r^j = \max\{f_r^i, f_r^j\}$, $r = 1,...,m$

### 2.2 S-Tree

S-tree [1, 3] is a tree with many branches that are balanced; each node of the S-tree contains a number of pairs $\langle sig, next\rangle$, where $sig$ is a binary signature and $next$ is a pointer to a child node. Each node root of the S-tree contains at least two pairs and at most $M$ pairs $\langle sig, next\rangle$, all internal nodes in the S-tree at least $m$ and at most $M$ pairs $\langle sig, next\rangle$, $1 \le m \le M/2$; the leaves of the S-tree contain the image's binary signatures $sig$, along with a unique identifier $oid$ for those images. The S-tree height for $n$ signatures is at most $h = \lceil \log_m n - 1 \rceil$. The S-tree was built on the basis of *inserting* and *splitting*. When the node $v$ is full, it will be split into two, at the same time the parent node $v_{parent}$ will created (if not exist) and two new signatures will be insert to node $v_{parent}$.

Each query signature will do the top-down order and can traverse many paths from root to leaf because the signature query can be match with many signatures at internal node in the S-tree.

### 2.3 FHD Distance

For two vector n-dimensional real value $x$ and $y$, difference fuzzy set as $D_\alpha(x, y)$, with membership function as $\mu_{D_\alpha(x,y)} = 1 - e^{-\alpha(x-y)^2}$. At that time the *fuzzy Hamming distance* [1, 2] between $x$ and $y$ with sign $FHD_\alpha(x, y)$ is the fuzzy cardinality of the difference fuzzy set $D_\alpha(x, y)$ and has membership function matching parameter $\alpha$ as $\mu_{FHD(x,y)}(\alpha) : \{0,1,...,n\} \to [0,1]$. Moreover $\mu_{FHD(x,y)}(k;\alpha) = \mu_{Card(D_\alpha(x,y))}(k)$, with $k \in \{0,1,...,n\}, n = |Support(D_\alpha(x, y))|$.

# 3 Building Data Structures and Image Retrieval Algorithms

## 3.1 Creating a Fuzzy Signature of the Color Image

*Step 1.* Choose a standard color set $C = \{c_1, c_2, ..., c_n\}$ to calculate the color histogram of the images. To quantify the image $I$ in order to retain only the dominant colors $C_I = \{c_1^I, c_2^I, ..., c_{n_I}^I\}$, the color histogram vector of image $I$ is $H_I = \{h_1^I, h_2^I, ..., h_{n_I}^I\}$.
*Step 2.* Calculate the color histogram vector standardizes $H = \{h_1, h_2, ..., h_n\}$, where $h_i = h_j^I \big/ \sum_j h_j^I$ if $c_i \in C \cap C_I$, otherwise $h_i = 0$.
*Step 3.* Each color $c_j^I$ will be described into a fuzzy signature with length m as $f_1^j f_2^j, ..., f_m^j$. Therefore, fuzzy signature of the color image $I$ will be $f_1^1 f_2^1, ..., f_m^1 f_1^2 f_2^2, ..., f_m^2 ... f_1^n f_2^n, ..., f_m^n$, in which: $f_i^j = \begin{cases} h_i & i = \lceil h_i \times m \rceil \\ 0 & i \neq \lceil h_i \times m \rceil \end{cases}$

Setting $F^j = f_1^j f_2^j ... f_m^j$, the fuzzy signature of the color image will be $FuzzySig = F^1 F^2 ... F^n$

## 3.2 The Similar Measure FHD

Each fuzzy signature $FuzzySig_I = F^1 F^2 ... F^n$ of image $I$ as vector $V_I = (v_1, v_2, ..., v_n)$, in which $v_i = weight(F^i) = \sum_{k=1}^{m} w_k^i$, with

$$F^i = f_1^i f_2^i ... f_m^i \quad \text{and} \quad w_k^i = \begin{cases} 0 & f_k^i = 0 \\ f_k^i + \dfrac{k}{m} \times 100 & f_k^i \neq 0 \end{cases}$$

Setting $J$ as an image to calculate the similarity in comparison with the image $I$, so we need to calculate the Hamming distance between two vector $V_I = (v_1^I, v_2^I, ..., v_n^I)$ and $V_J = (v_1^J, v_2^J, ..., v_n^J)$. The fuzzy distance FHD is as follows:

$$FHD_\alpha(V_I, V_J) = Card(D_\alpha(V_I, V_J)) = \sum_{i=0}^{n} i \big/ (\mu_{Card(D_\alpha(V_I, V_J))}(i))$$

In there, $\mu_{Card(D_\alpha(V_I, V_J))}(i) = \mu(i) \wedge (1 - \mu(i+1)) = \min\{\mu(i), (1 - \mu(i+1)\}$, $\mu(i)$ as an i-th largest value of function $\mu_i$ corresponds to fuzzy set $D_\alpha(V_I, V_J) = \sum_{i=1}^{n} i/\mu_i$ and $\mu(0) = 1, \mu(n+1) = 0$. At that time, the different levels of $V_I$ and $V_I$ on $k$ the component is as follows:
$\mu_{FHD_\alpha(V_I, V_J)}(k, \alpha) = \mu_{Card(D_\alpha(V_I, V_J))}(k)$, with $k \in \{0, 1, ..., n\}, n = |Support(D_\alpha(x, y))|$.

### 3.3 Creating S-Tree Based on FHD Distance

In order to reduce the storage space and increase the query speed, the paper aims to create the S-tree to store the fuzzy signatures of image. Each node in the S-tree stores elements $\{\langle FuzzySig, next\rangle\}$, with $FuzzySig$ is the fuzzy signature and $next$ is the reference pointer to child node. The leaf node store elements $\{\langle FuzzySig, Oid\rangle\}$, with $FuzzySig$ is a fuzzy signature of each image and $Oid$ is an identification of the corresponding image. The process of creating the S-tree is based on inserting and splitting the node in the tree [3, 6]. The algorithm which creates the S-tree to store the fuzzy signature is as follows:

```
Input: FS = {<FuzzySigi, Oidi> | i = 1,…,n}
Output: The S-tree
Algorithm1. Gen-FuzzyStree(S, Root)
Begin
  Step 1.  v = Root;
           If FS = ∅ then STOP;
           Else  Choosing <FuzzySig, Oid> ∈ S
                 S = S \ <FuzzySig, Oid>;
           To go step 2;
  Step 2.  If v is leaf then
           Begin
             v = v ∪ <FuzzySig, Oid>;
             UnionSignature(v);
             If v.count > M then SplitNode(v);
             To go back Step 1;
           End
           Else
           Begin
             FHD(SIG0→FuzzySig, Fuzzysig)
               = min{FDH(SIGi→FuzzySig,FuzzySig)|SIGi∈v};
             v = SIG0→next;
             To go back Step 2;
           End
End.
```

The ***Algorithm1*** gives signatures $FuzzySig$ from set of signatures $FS$ into the S-tree. Each signature $FuzzySig$ will be inserted into the suitable leaf node. If the leaf node is full, the process of splitting node will do and the S-tree will grow in height towards the root of the tree. At each inside node of the S-tree, the priority will go towards the similarity FHD more and this process will be approved until we can find a suitable leaf node.

Each signature need to insert will browse through the height $h = \lceil \log_m n - 1 \rceil$, with $m$ is the minimum signature of a node in S-tree. Setting $k$ is the length of each signature. Every node in tree has maximum $M$ signature. So, the browser to find the appropriate leaf node would cost a maximum of $k \times M \times \lceil \log_m n - 1 \rceil$. However, if we find the suitable node that is full, we have to split node. The splitting node relies on $\alpha - seed$, $\beta - seed$ is done as follows:

**Input:** Node v

**Output:** The S-tree after splitting node

***Algorithm2.*** SplitNode(v)

**Begin**

*Create the node* $v_\alpha$ *and* $v_\beta$ *contains* $\alpha - seed$ *and* $\beta - seed$

v = v \ { $\alpha - seed$ , $\beta - seed$ }

**For** ($SIG_i \in$ v)

**If** (FHD($SIG_i \to$FuzzySig, $\alpha - seed$ )<

EMD($SIG_i \to$Fuzzysig, $\beta - seed$ ))**then**

$v_\alpha = v_\alpha \cup SIG_i$;

**Else** $v_\beta = v_\beta \cup SIG_i$;

$s_\alpha = \vee SIG_i^\alpha$, with $SIG_i^\alpha \in v_\alpha$; $s_\beta = \vee SIG_i^\beta$, with $SIG_i^\beta \in v_\beta$;

**If** ( $v_{parent}$ != null) **then** $v_{parent} = v_{parent} \cup s_\alpha$; $v_{parent} = v_{parent} \cup s_\beta$;

**If** ( $v_{parent}.count$ > M ) **then** SplitNode( $v_{parent}$ );

**If** ( $v_{parent}$ = null) **then** Root = { $s_\alpha$ , $s_\beta$ };

**End.**

**Procedure** UnionSignature( $v$ )

**Begin**

$s = \vee SIG_i$ , with $SIG_i \in v$ ;

**If**( $v_{parent}$ != null) **then**

**Begin**

$SIG_v = \{SIG_i \mid SIG_i \to next = v, SIG_i \in v_{parent}\}$;

$v_{parent} \to (SIG_v \to FuzzySig) = s$;

UnionSignature( $v_{parent}$ );

**End**

**End.**

### 3.4 Image Retrieval Algorithm Based on the S-tree

After storing the signature and identification of image on the S-tree, the process of querying provides the signature of the image on the base of browsing the S-tree with the similarity measure FHD. After finding similar image signatures, based on the identification of specific image, we find the image similar to query image. For this reason, the problem is to find the signature of the image and identification of corresponding image. This is done as follows:

```
Input: Query signature FuzzySig and the S-tree
Output: SIGOUT={<SIGi, Oidi> } reference to images
Algorithm3. Search-Image-Sig(FuzzySig, S-tree)
Begin
  v = root; SIGOUT = ∅; Stack = ∅; Push(Stack, v);
  while(not Empty(Stack)) do
  Begin
  v = Pop(Stack);
  If(v is not Leaf) then
  begin
      For(SIGi∈v and SIGi→FuzzySig∧FuzzySig=FuzzySig) do
      FHD(SIG0→Fuzzysig, FuzzySig) =
  min{FHD(SIGi→FuzzySig, FuzzySig)| SIGi ∈ v};
      Push(Stack, SIG0 → next);
  end
  Else
      SIGOUT=SIGOUT∪{<SIGi→FuzzySig, Oidi>|SIGi ∈ v};
  end
  return SIGOUT;
End.
```

S-tree is a tree with many branches which are balanced, in addition to, at each node of the tree will be browsed in the direction of the next best similarity, which will cost up to browse the tree is $h = \lceil \log_m n - 1 \rceil$. The search process on the tree is done similar to the browsing of the tree, so the cost of the query on the S-tree will also be $k \times M \times \lceil \log_m n - 1 \rceil$, with $k$ is the length of each signature, $m$ is the minimum number of signature, $M$ is the maximum number of signature of one node on S-tree.

# 4 Experiment

## 4.1 Model Application

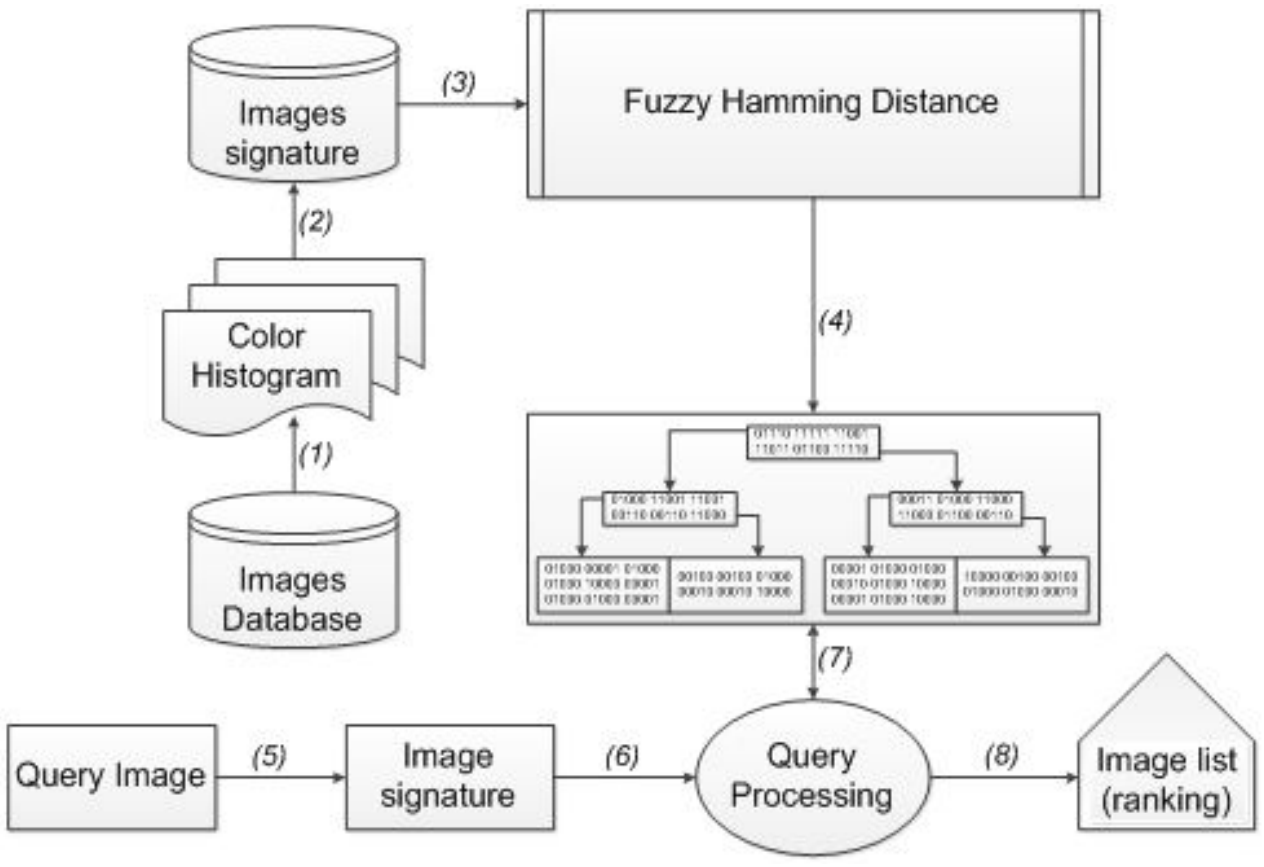

**Fig. 1.** Model image retrieval system using FHD and S-Tree

***Phase 1:*** *Perform Preprocessing*

*Step 1.* Quantize images in the database and convert to a color histogram.

*Step 2.* Convert the color histogram of the image in the form of fuzzy signatures.

*Step 3.* Respectively calculate the similarity measure FHD distance of the fuzzy signatures and insert into the S-tree.

***Phase 2:*** *Implementation Query*

*Step 1.* For each query image, calculate the color histogram and convert into fuzzy signatures.

*Step 2.* Perform fuzzy signature query on the S-tree consisting of the similar image signature at the leaves of the S-tree through the FHD measure.

*Step 3.* After finding similar images, conduct arrangement of similar levels from high to low and make the title match with the images arranged on the basis of similarity FHD distance.

## 4.2 The experimental results

Each image will calculate the color histogram based on 16 colors: BLACK, SILVER, WHITE, GRAY, RED, ORANGE, YELLOW, LIME, GREEN, TURQUOISE, CYAN, OCEAN, BLUE, VIOLET, MAGENTA, RASPBERRY.

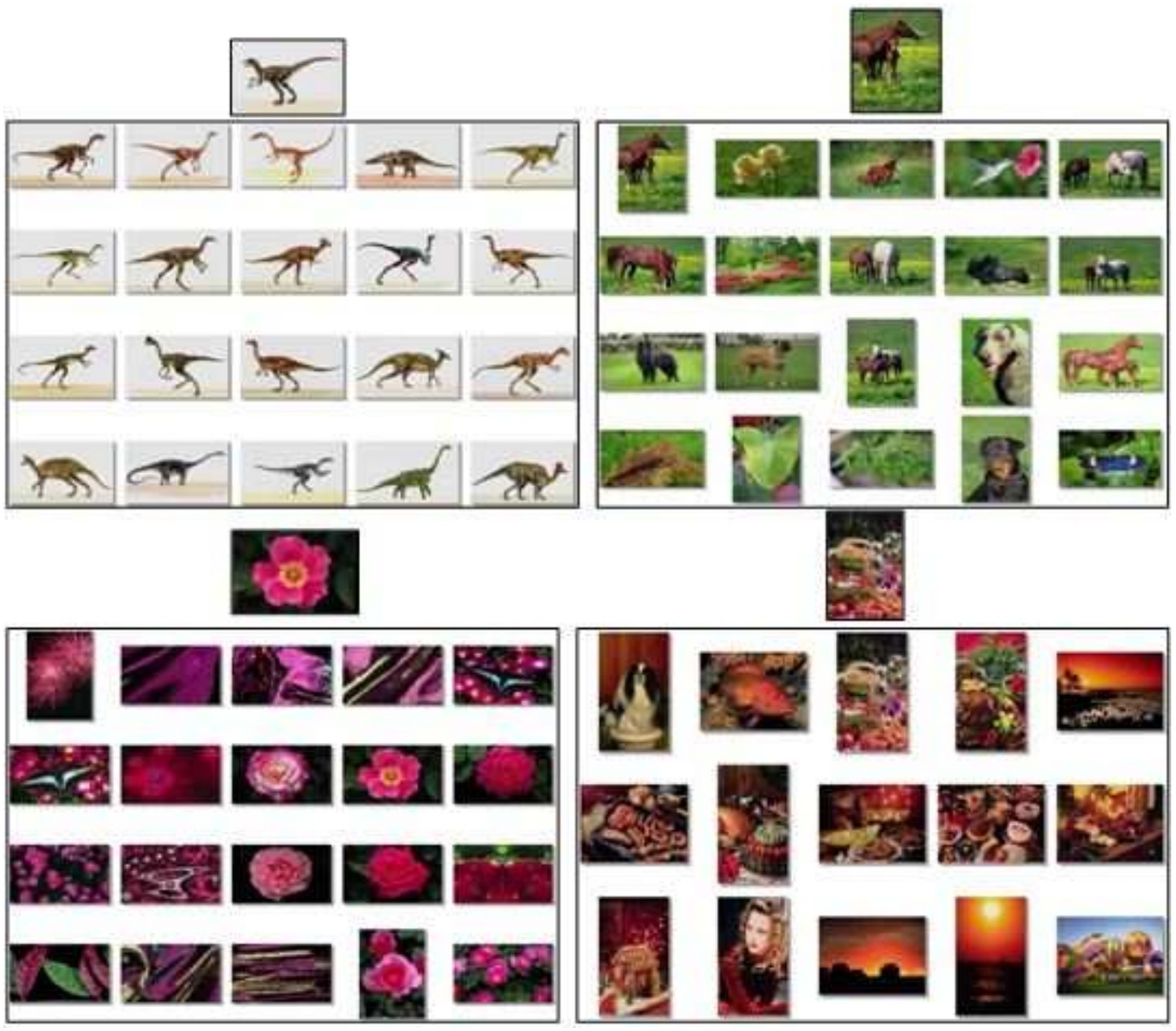

**Fig. 2.** Some results of the process query image in COREL database over 10,000 images

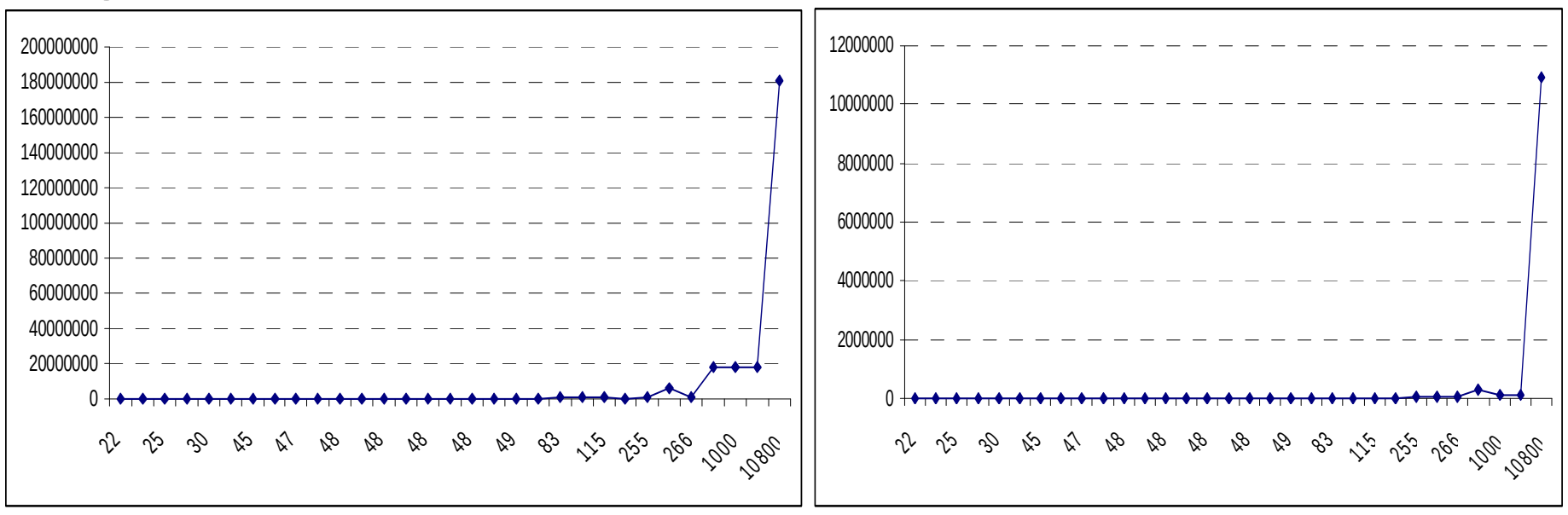

**Fig. 3.** Number of comparisons to create S-tree; the time to create S-tree (in milliseconds)

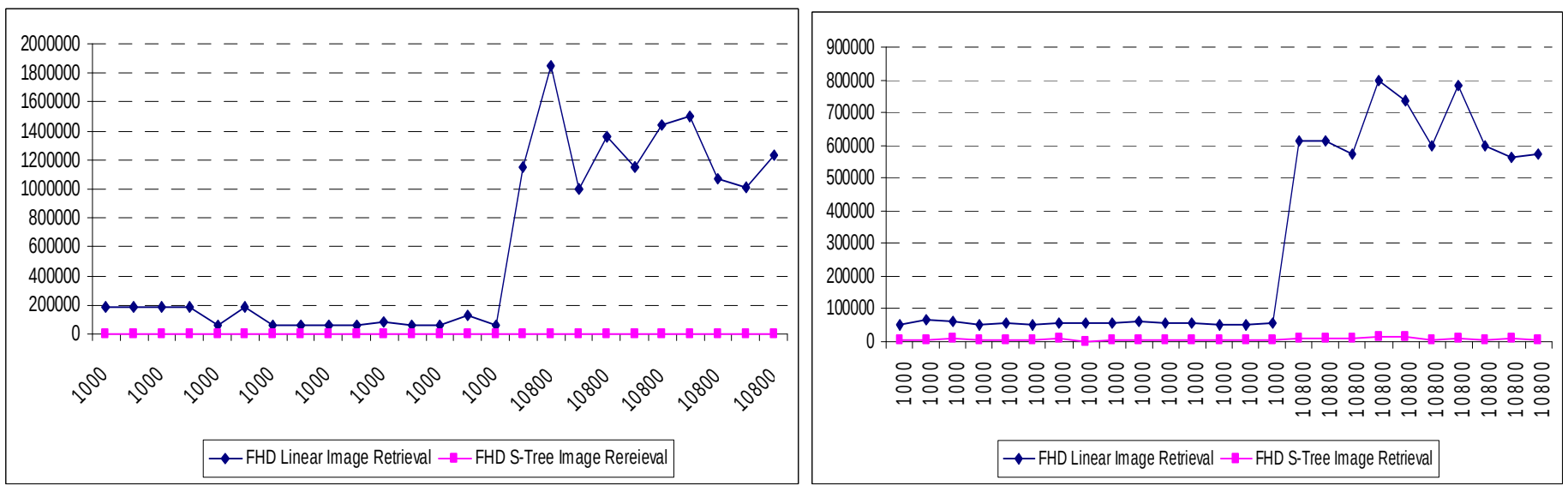

**Fig. 4.** Number of comparisons to query; the time to query (in milliseconds)

According to the experiment shows the process of creating the S-tree from the fuzzy signature of the image takes time-consuming, but the query image relies on the S-tree do much faster than the linear search method on the base of FHD distance.

## 5 Conclusion

In the experiment, the paper created the fuzzy signature to describe the color's image and showed the similar image retrieval algorithm, at the same time, experimented to query image on the base of the content. However, using the distribution of the image's color will result in inaccuracy in the case of images with the same percentage of color pixels, but the color distribution location does not correspond to each other. The next development will assess the similarity of the image with location distribution of the percentage of color pixels and compare objects in the contents of image to increase accuracy when querying the similar images.

## References

1. Vaclav Snasel, Zdenek Horak, Milos Kudelka, Ajith Abraham: Fuzzy Signatures Organized Using S-Tree, Proceedings of the IEEE International Conference on Digital Object Identifier, Anchorage, Alaska, 633--637, 9-12 Oct (2011)
2. Mircea M. Ionescu, Anca L. Ralescu: Image Clustering for a Fuzzy Hamming Distance Based CBIR System, Proceedings of the Sixteen Midwest Artificial Intelligence and Cognitive Science Conference, Dayton, 102--108, April (2005)
3. Yangjun Chen and Yibin Chen: On the Signature Tree Construction and Analysis, IEEE Trans. Knowl. Data Eng., 18(9), 1207--1224 (2006)
4. Neetu Sharma. S, Paresh Rawat. S, Jaikaran Singh. S.: Efficient CBIR Using Color Histogram Processing, Signal & Image Processing: An Inter. Jour., 2(1), 94--112 (2011)
5. Fazal Malik, Baharum Bin Baharudin: Feature Analysis of Quantized Histogram Color Features for Content-Based Image Retrieval Based on Laplacian Filter, International Conference on System Engineering and Modeling, 34, 44--49 (2012)
6. M. A. Nascimento, E. Tousidou, V. Chitkara, Y. Manolopoulos: Image indexing and retrieval using signature trees, Data & Knowledge Eng., 43(1), 57--77 (2002)
7. Rahul Mehta, Nishchol Mishra, Sanjeev Sharma, Color - Texture based Image Retrieval System, International Journal of Computer Applications, 24(5), 24--29 (2011)
8. Gunja Varshney, Uma Soni: Color-Based Image Retrieval in Image Database System, International Journal of Soft Computing and Engineering, 1(5), 31--35 (2011)